\documentclass[10pt,twocolumn,twoside]{IEEEtran} % for double.pdf

\usepackage{cite}
\usepackage{graphicx}
\ifCLASSINFOpdf
\else
\fi
\usepackage[cmex10]{amsmath}
\usepackage{algorithmic}
\usepackage{mdwmath}
\usepackage[tight,footnotesize]{subfigure}

\ifCLASSOPTIONcaptionsoff
  \usepackage[nomarkers]{endfloat}
 \let\MYoriglatexcaption\caption
 \renewcommand{\caption}[2][\relax]{\MYoriglatexcaption[#2]{#2}}
\fi
\usepackage{url}
\usepackage{multirow,algorithm}

\begin{document}

\title{Pairwise Constraint Propagation on Multi-View Data}


\author{Zhiwu~Lu~and~Liwei~Wang% <-this % stops a space
\thanks{Z. Lu is with the Key Laboratory of Data Engineering and Knowledge Engineering (MOE), School of Information, Renmin University of China, Beijing 100872, China (e-mail: zhiwu.lu@gmail.com). }
\thanks{L. Wang is with the Key Laboratory of Machine Perception (MOE), School of Electronics Engineering and Computer Science, Peking University, Beijing 100871, China (e-mail: wanglw@cis.pku.edu.cn). }
}

\maketitle

\begin{abstract}
This paper presents a graph-based learning approach to pairwise constraint propagation on multi-view data. Although pairwise constraint propagation has been studied extensively, pairwise constraints are usually defined over pairs of data points from a single view, i.e., only intra-view constraint propagation is considered for multi-view tasks. In fact, very little attention has been paid to inter-view constraint propagation, which is more challenging since pairwise constraints are now defined over pairs of data points from different views. In this paper, we propose to decompose the challenging inter-view constraint propagation problem into semi-supervised learning subproblems so that they can be efficiently solved based on graph-based label propagation. To the best of our knowledge, this is the first attempt to give an efficient solution to inter-view constraint propagation from a semi-supervised learning viewpoint. Moreover, since graph-based label propagation has been adopted for basic optimization, we develop two constrained graph construction methods for inter-view constraint propagation, which only differ in how the intra-view pairwise constraints are exploited. The experimental results in cross-view retrieval have shown the promising performance of our inter-view constraint propagation.
\end{abstract}



% \fi
\IEEEpeerreviewmaketitle

% main text
\section{Introduction}

As an alternative type of supervisory information easier to access
than the class labels of data points, pairwise constraints are widely
used for different machine learning tasks in the literature. To
effectively exploit pairwise constraints for clustering or classification \cite{lu2009generalized,lu2006iterative,wang2009image,lu2008comparing}, much attention has been paid to pairwise constraint propagation
\cite{LC08,LLT08,YS04}. Different from the method \cite{KKM03} which only adjusts the similarities between constrained
data points, these approaches can propagate pairwise constraints to other similarities between unconstrained data points and thus achieve better results in most cases. More importantly, given that each pairwise constraint is actually defined over a pair of data points from a single view, these approaches can all be regarded as intra-view constraint propagation when multi-view data is concerned. Since we have to learn the relationships (must-link or cannot-link) between data points, intra-view constraint propagation is more challenging than the traditional label propagation \cite{ZBLW04,ZGL03,lu2009image,lu2015noise,lu2011combining,lu2010combining} whose goal is only to predict the labels of unlabeled data points.

\begin{figure}[t]
\vspace{0.06in}
\begin{center}
\includegraphics[width=0.49\textwidth]{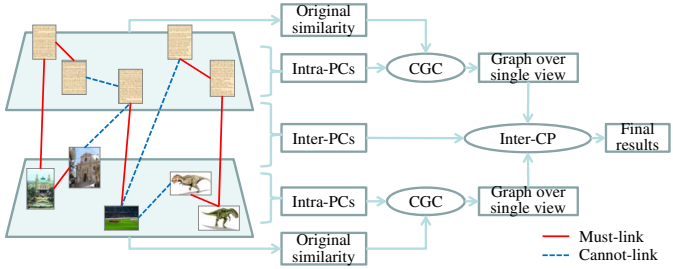}
\end{center}
\vspace{-0.15in} \caption{Illustration of the flowchart of inter-view constraint propagation (Inter-CP) with constrained graph construction (CGC). Here, we only consider two different views: text and image. Moreover, Intra-PCs and Inter-PCs denote intra-view
and inter-view pairwise constraints, respectively.} \label{Fig.1} \vspace{-0.15in}
\end{figure}

However, besides intra-view pairwise constraints, we may also have
easy access to inter-view pairwise constraints in multi-view tasks
such as cross-view retrieval \cite{RCC10}, where each pairwise
constraint is defined over a pair of data points from different views
(see Fig.~\ref{Fig.1}). In this case, inter-view pairwise
constraints still specify the must-link or cannot-link relationships
between data points. Since the similarity of two data points from
different views is commonly unknown in practice, inter-view
constraint propagation is significantly more challenging than
intra-view constraint propagation. In fact, very little attention has
been paid to inter-view constraint propagation for multi-view tasks
in the literature. Although pairwise constraint propagation has been
successfully applied to multi-view clustering in \cite{EDJ10,FIL11},
only intra-view pairwise constraints are propagated across different
views. Here, it should be noted that these two constraint propagation
methods \emph{have actually ignored} the concept of inter-view
pairwise constraints or the strategy of inter-view constraint
propagation.

Since multi-view data can be readily decomposed into a series of
two-view data, we focus on inter-view constraint propagation only
across two views in this paper. However, such inter-view constraint propagation remains a rather challenging task. Fortunately, from a
semi-supervised learning viewpoint, we can formulate inter-view constraint propagation as minimizing a regularized energy functional. Specifically, we first decompose the inter-view constraint propagation problem into a set of independent semi-supervised learning \cite{ZBLW04,ZGL03,lu2009image,lu2015noise} subproblems. Through formulating these subproblems uniformly as minimizing a regularized energy functional, we thus develop an efficient algorithm for inter-view constraint propagation based on the traditional graph-based label propagation technique \cite{ZBLW04}. In summary, we succeed in giving an insightful explanation of inter-view constraint propagation from a graph-based semi-supervised learning viewpoint.

However, since graph-based label propagation has been adopted for basic optimization, there remains one problem to be concerned in inter-view constraint propagation, i.e., how to exploit intra-view pairwise constraints for graph construction within each view. In this paper, we develop two constrained graph construction methods for inter-view constraint propagation, which only differ in how the intra-view pairwise constraints are exploited. The first method limits our inter-view constraint propagation to a single view and then utilize the constraint propagation results to adjust the weight matrix of each view, while the second method formulates graph construction as sparse representation and then directly add the intra-view pairwise constraints into sparse representation.

The flowchart of our inter-view constraint propagation with constrained graph construction is illustrated in Fig.~\ref{Fig.1}, where only two views (i.e. text and image) are considered. It should be noted that, when multiple views refer to text, image, audio and so on, the output of our inter-view constraint propagation actually denotes the correlation between different media views. That is, the proposed algorithm can be directly used for cross-view retrieval (also see examples in Fig.~\ref{Fig.3}) which has drawn much attention recently \cite{RCC10}. For cross-view retrieval, it is not feasible to combine multiple views just as previous multi-view retrieval methods \cite{GVS10,BMM08}. More notably, the two closely related methods \cite{EDJ10,FIL11} for multi-view clustering are actually incompetent for cross-view retrieval.

Finally, to emphasize our main contributions, we summarize the following distinct advantages of our pairwise constraint propagation on multi-view data:
\begin{itemize}
\item We have made the first attempt to give an efficient solution to inter-view constraint propagation from a graph-based semi-supervised learning viewpoint.
\item We have developed two constrained graph construction methods so that the intra-view pairwise constraints can also be exploited for inter-view constraint propagation.
\item When applied to cross-view retrieval, our inter-view constraint propagation has been shown to achieve promising results with respect to the state-of-the-art.
\item Although only evaluated in cross-view retrieval, our inter-view constraint propagation can be readily extended to many other multi-view tasks.
\end{itemize}

The remainder of this paper is organized as follows. In Section \ref{sect:intercp}, we formulate inter-view constraint propagation from a semi-supervised learning viewpoint. In Section \ref{sect:cgc}, we develop two constrained graph construction methods for our inter-view constraint propagation. In section \ref{sect:cmr}, our inter-view constraint propagation is applied to cross-view retrieval. Finally, Sections \ref{sect:exp} and  \ref{sect:con} provide the experimental results and conclusions, respectively.

\section{Inter-View Constraint Propagation}
\label{sect:intercp}

In this section, we first formulate inter-view constraint propagation
as minimizing a regularized energy functional from a semi-supervised
learning viewpoint. Furthermore, we develop an efficient algorithm for inter-view constraint propagation based on the label propagation technique \cite{ZBLW04}.

\subsection{Problem Formulation}

Given a set of inter-view pairwise constraints defined over pairs
of data points from different views, the goal of inter-view
constraint propagation is to learn the cross-view relationships
from these initial pairwise constraints. Since the similarity of two data points from different views is unknown in practice, inter-view
constraint propagation on multi-view data is much more challenging than the traditional pairwise constraint propagation over a single view. Considering that this multi-view problem can be readily decomposed into a series of two-view subproblems, we focus on inter-view constraint
propagation on two-view data in the following.

Let $\{\mathcal{X}, \mathcal{Y}\}$ be a two-view dataset, where $\mathcal{X} = \{x_1,...,x_N\}$ and $\mathcal{Y} = \{y_1,...,y_M\}$. It should be noted that we may have $N \neq M$. As an example, a two-view dataset is shown in Fig.~\ref{Fig.1}, with image and text being the two different views. For the two-view dataset $\{\mathcal{X}, \mathcal{Y}\}$, we can define a set of initial must-link constraints as $\mathcal{M} = \{(x_i,y_j): l(x_i)=l(y_j)\}$ and a set of initial cannot-link constraints as $\mathcal{C}=\{(x_i,y_j): l(x_i)\neq l(y_j)\}$, where $l(x_i)$ (or $l(y_j)$) is the class label of $x_i\in \mathcal{X}$ (or $y_j\in \mathcal{Y}$). Here, the two data points $x_i$ and $y_j$ are assumed to share the same class label set. If the class labels are not provided, the inter-view pairwise constraints can be defined only based on the correspondence between two views, which can be readily obtained from Web-based content (e.g. Wikipedia articles). Several examples of inter-view pairwise constraints are illustrated in Fig.~\ref{Fig.1}.

We can now state that the goal of inter-view constraint
propagation is to propagate the two sets of initial pairwise
constraints $\mathcal{M}$ and $\mathcal{C}$ across both
$\mathcal{X}$ and $\mathcal{Y}$. In fact, this is equivalent to
deriving the best solution $F^* \in \mathcal{F}$ from both
$\mathcal{M}$ and $\mathcal{C}$, with $\mathcal{F} = \{F =
{\{f_{ij}\}_{N\times M}}\}$. Here, any exhaustive set of
inter-view pairwise constraints is denoted as $F \in
\mathcal{F}$, where $f_{ij}>0$ means $(x_i,y_j)$ is a must-link
constraint while $f_{ij}<0$ means $(x_i,y_j)$ is a cannot-link
constraint, with $|f_{ij}|$ denoting the confidence score of
$(x_i,y_j)$ being a must-link (or cannot-link) constraint. Hence,
$\mathcal{F}$ can actually be regarded as the feasible solution set of inter-view constraint propagation.

\begin{figure}[t]
\vspace{0.06in}
\begin{center}
\includegraphics[width=0.30\textwidth]{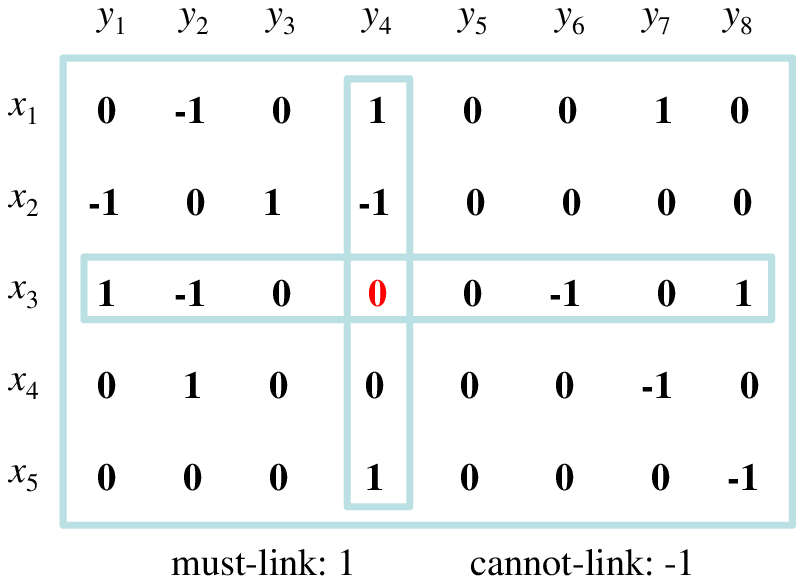}
\end{center}
\vspace{-0.15in}
\caption{Illustration of the initial matrix $Z$. When we focus on a
single pair of data points, e.g. $(x_3,y_4)$ here, the inter-view
constraint propagation can be viewed as a two-class semi-supervised
learning problem (in name only) in both vertical and horizontal
directions, where +1 (or -1) denotes positive (or negative) labeled
data and 0 denotes unlabeled data.} \label{Fig.2}
\vspace{-0.18in}
\end{figure}

Although it is difficult to directly find the best solution $F^* \in \mathcal{F}$ to inter-view constraint propagation, we can tackle
this challenging problem by decomposing it into a set of independent semi-supervised learning subproblems. More concretely, we first denote the two sets of initial pairwise constraints $\mathcal{M}$ and $\mathcal{C}$ with a single matrix $Z=\{z_{ij}\}_{N\times M}$:
\begin{eqnarray}
z_{ij} =\begin{cases}
+1, & (x_i,y_j)\in \mathcal{M}; \\
-1, & (x_i,y_j)\in \mathcal{C};\\
0, & \text{otherwise}.\\
\end{cases} \label{eq:pcs}
\end{eqnarray}
Moreover, by making vertical and horizontal observations on such
initial matrix $Z$, we decompose the inter-view constraint propagation
problem into independent semi-supervised learning subproblems, which is also illustrated in Fig.~\ref{Fig.2}. Finally, given two graphs $\mathcal{G}_{\mathcal{X}} = \{\mathcal{X}, W_{\mathcal{X}}\}$ and $\mathcal{G}_{\mathcal{Y}} = \{\mathcal{Y}, W_{\mathcal{Y}}\}$
constructed over $\{\mathcal{X}, \mathcal{Y}\}$ with $W_{\mathcal{X}}$ (or $W_{\mathcal{Y}}$) being the edge weight matrix defined over the vertex set $\mathcal{X}$ (or $\mathcal{Y}$), we utilize the graph-based label propagation method \cite{ZBLW04} to uniformly solve these semi-supervised learning subproblems:
\begin{eqnarray}
&\hspace{-0.3in}&\min_{F_\mathcal{X},F_\mathcal{Y}}\|F_\mathcal{X}-Z\|_{fro}^2 + \mu_\mathcal{X} \mathrm{tr}(F_\mathcal{X}^T\mathcal{L}_\mathcal{X}F_\mathcal{X}) + \|F_\mathcal{Y}-Z\|_{fro}^2 \nonumber \\
&\hspace{-0.3in}& \hspace{0.4in} + \mu_\mathcal{Y} \mathrm{tr} (F_\mathcal{Y}\mathcal{L}_\mathcal{Y}F_\mathcal{Y}^T) +
\gamma \|F_\mathcal{X}-F_\mathcal{Y}\|_{fro}^2,
\label{eq:intercp}
\end{eqnarray}
where $\mu_\mathcal{X}>0$ ($\mu_\mathcal{Y}>0$, or $\gamma>0$) denotes the regularization parameter, $\mathcal{L}_\mathcal{X}$ (or $\mathcal{L}_\mathcal{Y}$) denotes the normalized Laplacian matrix defined over $\mathcal{X}$ (or $\mathcal{Y}$), $||\cdot||_{fro}$ denotes the Frobenius norm of a matrix, and $\mathrm{tr}(\cdot)$ denotes the trace of a matrix.

The first and second terms of the above objective function are
related to the pairwise constraint propagation over $\mathcal{X}$, while the third and fourth terms are related to the pairwise constraint propagation over $\mathcal{Y}$. Moreover, the fifth term can ensure that the solutions of these two types of pairwise constraint propagation are as approximate as possible. Let $F_{\mathcal{X}}^*$ and $F_{\mathcal{Y}}^*$ be the best solutions of pairwise constraint propagation over $\mathcal{X}$ and $\mathcal{Y}$, respectively. The best solution of our inter-view constraint propagation is defined as follows:
\begin{eqnarray}
F^*=(F_{\mathcal{X}}^* + F_{\mathcal{Y}}^*)/2.
\end{eqnarray}
As for the second and fourth terms, they are known as the energy
functional \cite{ZGL03} (or smoothness) defined over
$\mathcal{X}$ and $\mathcal{Y}$. In summary, we have formulated intere-view constraint propagation as minimizing a regularized energy functional.

\subsection{Efficient Algorithm}
\label{subsect:alg}

Let $\mathcal{Q}(F_{\mathcal{X}},F_{\mathcal{Y}})$ denote the objective function in equation (\ref{eq:intercp}). The alternate optimization technique can be adopted to solve $\min_{F_{\mathcal{X}}, F_{\mathcal{Y}}} \mathcal{Q}(F_{\mathcal{X}},F_{\mathcal{Y}})$ as follows: 1) Fix $F_{\mathcal{Y}}=F_{\mathcal{Y}}^*$, and find $F_{\mathcal{X}}^*= \arg\min_{F_{\mathcal{X}}} \mathcal{Q}(F_{\mathcal{X}}, F_{\mathcal{Y}}^*)$; 2) Fix $F_{\mathcal{X}}=F_{\mathcal{X}}^*$, and find $F_{\mathcal{Y}}^*= \arg\min_{F_{\mathcal{Y}}} \mathcal{Q}(F_{\mathcal{X}}^*, F_{\mathcal{Y}})$.

\noindent \textbf{Pairwise Constraint Propagation over $\mathcal{X}$:} When $F_{\mathcal{Y}}$ is fixed at $F_{\mathcal{Y}}^*$, the solution of $\min_{F_{\mathcal{X}}}\mathcal{Q}(F_{\mathcal{X}},F_{\mathcal{Y}}^*)$ can be found by solving the following linear equation
\begin{eqnarray}
\frac{\partial \mathcal{Q}(F_{\mathcal{X}},F_{\mathcal{Y}}^*)}{2 \partial F_{\mathcal{X}}} = (F_{\mathcal{X}}-Z) + \mu_{\mathcal{X}} \mathcal{L}_{\mathcal{X}} F_{\mathcal{X}} + \gamma(F_{\mathcal{X}} -F_{\mathcal{Y}}^*)= 0,\nonumber
\end{eqnarray}
which can be equivalently transformed into:
\begin{eqnarray}
(I+\hat{\mu}_{\mathcal{X}}\mathcal{L}_{\mathcal{X}})F_{\mathcal{X}} = (1-\beta)Z+\beta F_{\mathcal{Y}}^*, \label{eq:lex}
\end{eqnarray}
where $\hat{\mu}_{\mathcal{X}}= \mu_{\mathcal{X}}/(1+\gamma)$ and $\beta=\gamma/(1+\gamma)$. Since $I+\hat{\mu}_{\mathcal{X}} \mathcal{L}_{\mathcal{X}}$ is positive definite, we then obtain an analytical solution:
\begin{eqnarray}
F_{\mathcal{X}}^* = (I+ \hat{\mu}_{\mathcal{X}} \mathcal{L}_{\mathcal{X}})^{-1}((1-\beta)Z+\beta F_{\mathcal{Y}}^*). \label{eq:lpx}
\end{eqnarray}
However, this analytical solution is not efficient for large
datasets, since matrix inverse has a time cost of $O(N^3)$.
Fortunately, equation (\ref{eq:lex}) can also be \emph{efficiently found using label propagation} \cite{ZBLW04} with $k$-nearest neighbor ($k$-NN) graph.

\noindent \textbf{Pairwise Constraint Propagation over $\mathcal{Y}$:} When $F_{\mathcal{X}}$ is fixed at $F_{\mathcal{X}}^*$, the solution of $\min_{F_{\mathcal{Y}}}\mathcal{Q}(F_{\mathcal{X}}^*,F_{\mathcal{Y}})$ can be found by solving the following linear equation
\begin{eqnarray}
\frac{\partial \mathcal{Q}(F_{\mathcal{X}}^*,F_{\mathcal{Y}})}{2 \partial F_{\mathcal{Y}}} = (F_{\mathcal{Y}}-Z) + \mu_{\mathcal{Y}} F_{\mathcal{Y}}\mathcal{L}_{\mathcal{Y}} + \gamma(F_{\mathcal{Y}}-F_{\mathcal{X}}^*)= 0,\nonumber
\end{eqnarray}
which can be equivalently transformed into:
\begin{eqnarray}
F_{\mathcal{Y}}(I+\hat{\mu}_{\mathcal{Y}}\mathcal{L}_{\mathcal{Y}}) = (1-\beta)Z+\beta F_{\mathcal{X}}^*,\label{eq:ley}
\end{eqnarray}
where $\hat{\mu}_{\mathcal{Y}}= \mu_{\mathcal{Y}}/(1+\gamma)$ and $\beta=\gamma/(1+\gamma)$. Since $I+\hat{\mu}_{\mathcal{Y}} \mathcal{L}_{\mathcal{Y}}$ is positive definite, we then obtain an analytical solution:
\begin{eqnarray}
F_{\mathcal{Y}}^* = ((1-\beta)Z+\beta F_{\mathcal{X}}^*) (I+\hat{\mu}_{\mathcal{Y}}\mathcal{L}_{\mathcal{Y}})^{-1}, \label{eq:lpy}
\end{eqnarray}
which involves time-consuming matrix inverse. In fact, the linear equation (\ref{eq:ley}) can also be efficiently solved using label propagation \cite{ZBLW04} with $k$-NN graph.

Let $W_\mathcal{X}$ (or $W_\mathcal{Y}$) denote the weight matrix of the $k$-NN graph constructed over $\mathcal{X}$ (or $\mathcal{Y}$). The complete algorithm for inter-view constraint propagation is summarized as follows:
\begin{description}
\item[(1)] Compute two matrices $S_\mathcal{X} =
    D_{\mathcal{X}}^{-1/2} W_\mathcal{X} D_{\mathcal{X}}^{-1/2}$
    and $S_\mathcal{Y} = D_{\mathcal{Y}}^{-1/2} W_\mathcal{Y}
    D_{\mathcal{Y}}^{-1/2}$, where $D_\mathcal{X}$ (or
    $D_\mathcal{Y}$) is a diagonal matrix with its $i$-th
    diagonal entry being the sum of the $i$-th row of
    $W_\mathcal{X}$ (or $W_\mathcal{Y}$);
\item[(2)] Initialize $F_{\mathcal{X}}(0)=0$,
    $F_{\mathcal{Y}}^*=0$, and $F_{\mathcal{Y}}(0)=0$;
\item[(3)] Iterate $F_{\mathcal{X}}(t+1) = \alpha_{\mathcal{X}}
    S_{\mathcal{X}} F_{\mathcal{X}}(t) +
    (1-\alpha_{\mathcal{X}})((1-\beta)Z+\beta F_{\mathcal{Y}}^*)$
    until convergence at $F_{\mathcal{X}}^*$, where
    $\alpha_{\mathcal{X}}=\hat{\mu}_{\mathcal{X}}/(1+\hat{\mu}_{\mathcal{X}})$    and $\beta=\gamma/(1+\gamma)$;
\item[(4)] Iterate $F_{\mathcal{Y}}(t+1) = \alpha_{\mathcal{Y}}
    F_{\mathcal{Y}}(t)S_{\mathcal{Y}} +
    (1-\alpha_{\mathcal{Y}})((1-\beta)Z+\beta F_{\mathcal{X}}^*)$
    until convergence at $F_{\mathcal{Y}}^*$, where
    $\alpha_{\mathcal{Y}}=\hat{\mu}_{\mathcal{Y}}/(1+\hat{\mu}_{\mathcal{Y}})$;
\item[(5)] Iterate Steps (3)--(4) until convergence, and output the final solution $F^*=(F_{\mathcal{X}}^*+ F_{\mathcal{Y}}^*)/2$.
\end{description}

According to the convergence analysis in \cite{ZBLW04}, Step (3) converges to $ F_{\mathcal{X}}^*=(1-\alpha) (I-\alpha_{\mathcal{X}} S_{\mathcal{X}})^{-1} ((1-\beta)Z+\beta F_{\mathcal{Y}}^*)$, equal to the solution (\ref{eq:lpx}) given that $\alpha_{\mathcal{X}}= \hat{\mu}_{\mathcal{X}}/(1+ \hat{\mu}_{\mathcal{X}})$ and $S_{\mathcal{X}}=I- \mathcal{L}_{\mathcal{X}}$. Similarly, Step (4) converges to $ F_{\mathcal{Y}}^*=(1-\alpha)((1-\beta)Z+\beta F_{\mathcal{X}}^*)(I-\alpha_{\mathcal{Y}} S_{\mathcal{Y}})^{-1}$, equal to the solution (\ref{eq:lpy}) given that $\alpha_{\mathcal{Y}}= \hat{\mu}_{\mathcal{Y}}/(1+ \hat{\mu}_{\mathcal{Y}})$ and $S_{\mathcal{Y}}=I- \mathcal{L}_{\mathcal{Y}}$. In the experiments, we find that Steps (3)--(5) generally converge in very limited iterations ($<$10). Moreover, based on $k$-NN graphs, the above inter-view constraint propagation algorithm has a time cost of $O(kNM)$, which is proportional to the number of all possible inter-view pairwise constraints. Hence, we consider that this algorithm can provide an efficient solution to inter-view constraint propagation (note that even a simple assignment operator on $F^*$ incurs a time cost of $O(NM)$).

\section{Constrained Graph Construction}
\label{sect:cgc}

In the last section, we have just developed an efficient inter-view constraint propagation algorithm based on the graph-based label propagation technique. However, since graph-based label propagation has been adopted as a basic optimization technique, there remains one problem to be concerned in inter-view constraint propagation, i.e., how to exploit intra-view pairwise constraints for graph construction within each view. In this section, we then develop two constrained graph construction methods for inter-view constraint propagation, which only differ in how the intra-view pairwise constraints are exploited. To ensure our inter-view constraint propagation algorithm runs efficiently even on large datasets, we utilize the traditional $k$-NN graph construction as the basis of our constrained graph construction, i.e., the obtained two constrained graphs can be considered as the variants of $k$-NN graph. In the following, we will only elaborate how to construct the graph $\mathcal{G}_{\mathcal{X}} = \{\mathcal{X}, W_{\mathcal{X}}\}$ over $\mathcal{X}$. The graph $\mathcal{G}_{\mathcal{Y}} = \{\mathcal{Y}, W_{\mathcal{Y}}\}$ over $\mathcal{Y}$ can be constructed exactly in the same way.

\subsection{Constrained Weight Adjustment}
\label{subsect:csa}

 The first constrained graph construction method limits our inter-view constraint propagation proposed in Section~\ref{sect:intercp} to a single view (i.e. intra-view constraint propagation over $\mathcal{X}$) and then utilize the obtained results of intra-view constraint propagation to adjust the weight matrix, which is thus called as constrained weight adjustment (CWA). According to the convergence analysis in Section~\ref{subsect:alg}, we construct a $k$-NN graph over $\mathcal{X}$ to speed up our intra-view constraint propagation.

\subsubsection{Intra-View Constraint Propagation}

We have just provided a sound solution to the challenging problem of
intra-view constraint propagation in Section~\ref{sect:intercp}. In this subsection, we further consider pairwise constraint propagation over a single view, where each pairwise constraint is defined over a pair of data points from the same view. In fact, this intra-view constraint propagation problem can also be solved from a semi-supervised learning viewpoint by limiting our inter-view constraint propagation to a single view.

Given the dataset $\mathcal{X}=\{x_1,...,x_N\}$, we
denote the set of initial must-link constraints as $\mathcal{M}_{\mathcal{X}} = \{(x_i,x_j): l_i=l_j\}$ and the set of initial cannot-link constraints as $\mathcal{C}_{\mathcal{X}} = \{(x_i,x_j): l_i\neq l_j\}$, where $l_i$ is the label of data point $x_i$. Similar to our representation of the initial inter-view pairwise constraints, we first denote the initial intra-view pairwise constraints $\mathcal{M}_{\mathcal{X}}$ and $\mathcal{C}_{\mathcal{X}} $ with a single matrix $Z_{\mathcal{X}}=\{z^{(x)}_{ij}\}_{N\times N}$:
\begin{eqnarray}
z^{(x)}_{ij} =\begin{cases}
+1, & (x_i,x_j)\in \mathcal{M}_{\mathcal{X}}; \\
-1, & (x_i,x_j)\in \mathcal{C}_{\mathcal{X}};\\
0, & \text{otherwise}.\\
\end{cases} \label{eq:pcs}
\end{eqnarray}
Furthermore, by making vertical and horizontal observations on $Z_{\mathcal{X}}$, we further decompose the intra-view constraint propagation problem into semi-supervised learning subproblems, just as our interpretation of inter-view constraint propagation from a semi-supervised learning viewpoint. These subproblems can be similarly merged to a single optimization problem (similar to \cite{LI10,fu2011symmetric,lu2013exhaustive}):
\begin{eqnarray}
&\hspace{-0.3in}&\min_{F_v,F_h}\|F_v-Z_{\mathcal{X}}\|_{fro}^2 +
\mu \mathrm{tr}(F_v^T\mathcal{L}_{\mathcal{X}} F_v) + \|F_h-Z_{\mathcal{X}}\|_{fro}^2 \nonumber \\
&\hspace{-0.3in}& \hspace{0.5in} +\mu \mathrm{tr}(F_h\mathcal{L}_{\mathcal{X}} F_h^T) + \gamma \|F_v-F_h\|_{fro}^2,
\label{eq:intracp}
\end{eqnarray}
where $\mu>0$ (or $\gamma>0$) denotes the regularization parameter, and $\mathcal{L}_{\mathcal{X}}$ denotes the normalized Laplacian matrix defined over the $k$-NN graph. The second and fourth terms of the above equation denote the energy functional \cite{ZGL03} (or the smoothness measure) defined over $\mathcal{X}$. In summary, we have also formulated intra-view constraint propagation as minimizing a regularized energy functional.

Similar to what we have done for solving equation (\ref{eq:intercp}), we can adopt the alternate optimization technique to find the best solution to the above intra-view constraint propagation problem. Let $W_{\mathcal{X}}$ denote the weight matrix of the $k$-NN graph constructed over the dataset $\mathcal{X}$. The proposed algorithm for our intra-view constraint propagation is outlined as follows:
\begin{description}
\item[(1)] Compute $S_{\mathcal{X}} = D^{-\frac{1}{2}} W_{\mathcal{X}} D^{-\frac{1}{2}}$, where $D$ is a diagonal matrix with its entry $(i,i)$ being the sum
    of row $i$ of $W_{\mathcal{X}}$;
\item[(2)] Initialize $F_v(0)=0$, $F_h^*=0$, and $F_h(0)=0$;
\item[(3)] Iterate $F_v(t+1) = \alpha S_{\mathcal{X}} F_v(t)
    +(1-\alpha)((1-\beta)Z_{\mathcal{X}}+\beta F_h^*)$ until convergence at $F_v^*$, where $\alpha=\mu/(1+\mu+\gamma)$ and $\beta= \gamma/(1+\gamma)$;
\item[(4)] Iterate $F_h(t+1) = \alpha F_h(t) S_{\mathcal{X}}
    +(1-\alpha)((1-\beta)Z_{\mathcal{X}} +\beta F_v^*)$ until convergence at $F_h^*$;
\item[(5)] Iterate Steps (3)--(4) until the stopping condition is
    satisfied, and obtain $F^*= (F_v^* + F_h^*)/2$.
\item[(6)] Output the normalized solution $F^*=F^*/F^*_{max}$, where $F^*_{max}$ denotes the maximum entry of $F^*$.
\end{description}
In the experiments, we find that Steps (3)--(5) generally converge in very limited iterations ($<$10). Moreover, based on $k$-NN graph, our algorithm has a time cost of $O(kN^2)$ proportional to the number of all possible pairwise constraints. Hence, it can be
considered to provide an efficient solution.

\subsubsection{Weight Adjustment Using Propagated Constraints}

It should be noted that the normalized output $F^*= \{f^*_{ij}\}_{N\times N}$ of our intra-view constraint propagation represents an exhaustive set of intra-view pairwise constraints. Our original motivation is to construct a new graph over $\mathcal{X}$ that is fully consistent with $F^*$. In fact, we can exploit $F^*$ for such graph construction by adjusting the original normalized weight matrix $W_{\mathcal{X}}$ (i.e. $0 \leq w^{(x)}_{ij} \leq 1$) just as \cite{LI10}:
\begin{eqnarray}
\tilde{w}^{(x)}_{ij} =\begin{cases}
1-(1-f^*_{ij})(1-w^{(x)}_{ij}), & f^*_{ij}\geq 0; \\
(1+f^*_{ij})w^{(x)}_{ij}, & f^*_{ij} < 0.
\end{cases} \label{eq:simadjust}
\end{eqnarray}
Since $\tilde{W}_{\mathcal{X}}=\{\tilde{w}^{(x)}_{ij}\}_{N\times N}$  is nonnegative and symmetric, we then use it as the new weight matrix. Moreover, we can find that $\tilde{W}^{(x)}_{ij} \geq W^{(x)}_{ij}$ (or $< W^{(x)}_{ij}$) if $F^*_{ij} \geq 0$ (or $< 0$). That is, the new weight matrix $\tilde{W}_{\mathcal{X}}$ is derived from the original weight matrix $W_{\mathcal{X}}$ by increasing $W^{(x)}_{ij}$ for the must-link constraints with $F^*_{ij}>0$ and decreasing $W^{(x)}_{ij} $ for the cannot-link constraints with $F^*_{ij}<0$. This is entirely consistent with our original motivation of exploiting intra-view
pairwise constraints for graph construction.

Once we have constructed the new weight matrix $\tilde{W}_{\mathcal{X}}$ over $\mathcal{X}$, we can similarly construct the new weight matrix $\tilde{W}_{\mathcal{Y}}$ over $\mathcal{Y}$. Based on these two new weight matrices, our inter-view constraint propagation can be performed with constrained graph construction (CGC) (as shown in Fig.~\ref{Fig.1}) using constrained weight adjustment (CWA) developed here.

\subsection{Constrained Sparse Representation}
\label{subsect:csr}

The second constrained graph construction method formulates graph construction as sparse representation \cite{Donoho04,WYG09,lu2011latent} and then directly add the intra-view pairwise constraints into sparse representation, which is thus called as constrained sparse representation (CSR). Our work is mainly inspired by recent effort to exploit sparse representation for graph construction, i.e., $L_1$-graph construction \cite{CLY09,CYS10}. The basic idea of $L_1$-graph construction is to seek a sparse linear reconstruction of each data point with the other data points. However, such $L_1$-graph construction may become infeasible since it incurs too much time cost given a large data size $N$. Hence, we only consider the $k$ nearest neighbors of each data point for its sparse linear reconstruction, which thus becomes a much smaller scale optimization problem ($k \ll N$). More notably, due to such neighborhood limitation, the obtained $L_1$-graph is actually a variant of $k$-NN graph, which can ensure that our inter-view constraint propagation proposed in Section~\ref{sect:intercp} runs efficiently on large datasets. Finally, to exploit intra-view pairwise constraints for $L_1$-graph construction, we seek a constrained sparse linear reconstruction of each data point.

\subsubsection{$L_1$-Graph Construction with Sparse Representation}

We start with the problem formulation for sparse linear
reconstruction of each data point in its $k$-nearest neighborhood.
Given a data point $x_i \in \mathcal{X}$, we suppose it can be
reconstructed using its $k$-nearest neighbors (their indices are
collected into $\mathcal{N}_k(i)$), which results in an
underdetermined linear system: $x_i = B_i \alpha_i$, where $\alpha_i
\in R^k$ is a vector that stores unknown reconstruction
coefficients, and $B_i = [x_j]_{j\in \mathcal{N}_k(i)}$ is an
overcomplete dictionary with $k$ bases. According to
\cite{Donoho04}, if the solution for $x_i$ is sparse enough, it can
be recovered by:
\begin{eqnarray}
\min_{\alpha_i}~~||\alpha_i||_1,~~\mathrm{s.t.}~~x_i = B_i \alpha_i,
\end{eqnarray}
where $||\alpha_i||_1$ is the $L_1$-norm of $\alpha_i$. Given the
kernel (affinity) matrix $A=\{a_{ij}\}_{N\times N}$ computed over
$\mathcal{X}$, we make use of the kernel trick and transform the
above problem into:
\begin{eqnarray}
\min_{\alpha_i}~~||\alpha_i||_1,~~\mathrm{s.t.}~~ \hat{x}_{i} = C_i
\alpha_i, \label{eq:sr}
\end{eqnarray}
where $\hat{x}_i=[a_{ji}]_{j\in \mathcal{N}_k(i)} \in R^k$,
$C_i=[a_{jj'}]_{j,j'\in \mathcal{N}_k(i)} \in R^{k\times k}$. In
practice, due to the noise in the data, we can reconstruct $\hat{x}_i$ similar to \cite{WYG09}: $\hat{x}_i = C_i \alpha_i + \zeta_i$, where
$\zeta_i$ is the noise term. The above $L_1$-optimization problem
can then be redefined by minimizing the $L_1$-norm of both
reconstruction coefficients and reconstruction error:
\begin{eqnarray}
\min_{\alpha'_i}~~||\alpha'_i||_1,~~\mathrm{s.t.}~~\hat{x}_i = C'_i
\alpha'_i, \label{eq:srsol}
\end{eqnarray}
where $C'_i = [C_i, I] \in R^{k \times 2k}$ and $\alpha'_i =
[\alpha^T_i, \zeta^T_i ]^T$. This convex optimization can be solved
by general linear programming and has a globally optimal solution.

After we have obtained the reconstruction coefficients for all the
data points by the above sparse linear reconstruction, the weight
matrix $W_{\mathcal{X}}=\{w^{(x)}_{ij}\}_{N\times N}$ can be defined
by:
\begin{eqnarray}
w^{(x)}_{ij} =\begin{cases}
|\alpha'_i(j')|,   & j\in\mathcal{N}_k(i),j'=\mathrm{index}(j,\mathcal{N}_k(i)); \\
0,   & \mathrm{otherwise}, \\
\end{cases}  \label{eq:l1wt}
\end{eqnarray}
where $\alpha'_i(j')$ denotes the $j'$-th element of the vector
$\alpha'_i$, and $j'=\mathrm{index}(j,\mathcal{N}_k(i))$ means that
$j$ is the $j'$-th element of the set $\mathcal{N}_k(i)$. By setting
the weight matrix $W_{\mathcal{X}} = (W_{\mathcal{X}} +
W_{\mathcal{X}}^T)/2$, we construct a graph
$\mathcal{G}_{\mathcal{X}} = \{\mathcal{X}, W_{\mathcal{X}}\}$ over
$\mathcal{X}$, which is called as $L_1$-graph since it is
constructed by $L_1$-optimization.

\subsubsection{$L_1$-Norm Laplacian Regularization with Intra-View Pairwise Constraints}

In the above $L_1$-graph construction, we have ignored intra-view
pairwise constraints (see examples in Fig.~\ref{Fig.1}).
In fact, this supervisory information can be exploited for
$L_1$-graph construction through Laplacian regularization
\cite{ZBLW04,ZGL03}. Our basic idea is to first derive
Laplacian regularization from intra-view pairwise constraints and
then incorporate this constrained term into sparse linear
reconstruction (the key step of $L_1$-graph construction).
In the following, we will first elaborate how to derive a new
Laplacian regularization term from intra-view pairwise constraints.

Given a set of intra-view must-link constraints
$\mathcal{M}_\mathcal{X}$ and a set of intra-view cannot-link
constraints $\mathcal{C}_\mathcal{X}$ defined over $\mathcal{X}$, we
can represent both $\mathcal{M}_\mathcal{X}$ and
$\mathcal{C}_\mathcal{X}$ using a single matrix
$Z_\mathcal{X}=\{z^{(x)}_{ij}\}_{N\times N}$ exactly the same as equation (\ref{eq:pcs}). The normalized Laplacian matrix limited to the $k$-nearest neighborhood of data point $x_i$ can thus be defined as:
\begin{eqnarray}
\mathcal{L}_i = I-D_i^{-1/2}(1+Z_i)D_i^{-1/2},
\end{eqnarray}
where $Z_i=[z^{(x)}_{jj'}]_{j,j'\in \mathcal{N}_k(i)}\in R^{k\times
k}$, and $D_i$ is a diagonal matrix with its $j$-th diagonal element
being the sum of the $j$-th row of $1+Z_i$. Here, we define the
similarity matrix (i.e. $1+Z_i$) limited to the $k$-nearest
neighborhood $\mathcal{N}_k(i)$ of $x_i$ based on the intra-view
pairwise constraints stored in $Z_\mathcal{X}$. From this normalized
Laplacian matrix $\mathcal{L}_i$, we can derive the Laplacian
regularization term for the sparse representation problem
(\ref{eq:sr}) as $\alpha_i^T\mathcal{L}_i \alpha_i$, the same as the original definition in \cite{ZBLW04}.

However, we have difficulty in directly incorporating this Laplacian
regularization term into the sparse representation problem
(\ref{eq:sr}), no matter as a part of the objective function or a
constraint condition. Hence, we further formulate an $L_1$-norm
version of Laplacian regularization \cite{lu2012image,lu2013latent,lu2015noise,lu2015semantic}:
\begin{eqnarray}
||\tilde{C}_i  \alpha_i||_1 =
||\Sigma_i^{\frac{1}{2}}V_i^T\alpha_i||_1,
\end{eqnarray}
where $\tilde{C}_i=\Sigma_i^{\frac{1}{2}}V_i^T$, $V_i$ is a $k\times
k$ orthonormal matrix with each column being an eigenvector of
$\mathcal{L}_i$, and $\Sigma_i$ is a $k \times k$ diagonal matrix
with its diagonal element $\Sigma_i(j,j)$ being an eigenvalue of
$\mathcal{L}_i$ (sorted as $\Sigma_i(1,1) \leq ...\leq
\Sigma_i(k,k)$). Given that $\mathcal{L}_i$ is nonnegative definite,
$\Sigma_i \geq 0$ (i.e. all the eigenvalues $\geq 0$). Since
$\mathcal{L}_i V_i = V_i \Sigma_i$ and $V_i$ is orthonormal, we have
$\mathcal{L}_i = V_i \Sigma_i V_i^T$. Hence, the original Laplacian
regularization $\alpha_i^T\mathcal{L}_i \alpha_i$ can be
reformulated as:
\begin{eqnarray}
\alpha_i^T\mathcal{L}_i \alpha_i = \alpha_i^T V_i
\Sigma_i^{\frac{1}{2}} \Sigma_i^{\frac{1}{2}} V_i^T \alpha_i =
||\tilde{C}_i \alpha_i||_2^2,
\end{eqnarray}
which means that our new formulation $||\tilde{C}_i \alpha_i||_1$
can indeed be regarded as an $L_1$-norm version of the original
Laplacian regularization $\alpha_i^T\mathcal{L}_i \alpha_i=
||\tilde{C}_i \alpha_i||_2^2$.

\subsubsection{$L_1$-Graph Construction with $L_1$-Norm Laplacian Regularization}

After we have formulated  $L_1$-norm Laplacian regularization based
on intra-view pairwise constraints, we can further incorporate this
constrained term into sparse linear reconstruction used for
$L_1$-graph construction. More concretely, by introducing noise
terms for linear reconstruction and $L_1$-norm Laplacian
regularization, we transform the sparse representation problem
(\ref{eq:sr}) into
\begin{eqnarray}
&&\min_{\alpha_i,\zeta_i,\xi_i}~~||[\alpha_i^T,\zeta_i^T,
\xi_i^T]||_1, \nonumber \\
&&\mathrm{s.t.}~~\hat{x}_i = C_i\alpha_i+\zeta_i,~0=\tilde{C}_i
\alpha_i+\xi_i, \label{eq:ssr}
\end{eqnarray}
where the reconstruction error and Laplacian regularization with
respect to $\alpha_i$ are controlled by $\zeta_i$ and $\xi_i$,
respectively. Let $\alpha'_i=[\alpha_i^T,\zeta_i^T, \xi_i^T]^T$,
$C'_i=\left[
  \begin{array}{ccc}
    C_i & I & 0 \\
    \tilde{C}_i & 0 & I \\
  \end{array}
\right]$, and $\hat{x}'_i=[\hat{x}_i^T,0^T]^T$. We finally solve the
following constrained spare representation problem for $L_1$-graph
construction:
\begin{eqnarray}
\min_{\alpha'_i}~~||\alpha'_i||_1,~~\mathrm{s.t.}~~\hat{x}'_i = C'_i
\alpha'_i,
\end{eqnarray}
which takes the same form as the original spare representation
problem (\ref{eq:srsol}). Here, it is noteworthy that this constrained
spare representation (CSR) problem can be solved very efficiently, since it is limited to $k$-nearest neighborhood. The weight matrix
$W_{\mathcal{X}}$ of the $L_1$-graph $\mathcal{G}_{\mathcal{X}} =
\{\mathcal{X}, W_{\mathcal{X}}\}$ can be defined the same as
equation (\ref{eq:l1wt}).

In our CSR formulation, the $L_1$-norm Laplacian regularization can be smoothly incorporated into the original sparse representation problem (\ref{eq:sr}). However, this is not true for the traditional Laplacian regularization \cite{ZBLW04,ZGL03}, which may introduce extra parameters (hard to tune in practice) into the $L_1$-optimization for sparse representation. Meanwhile, our $L_1$-norm Laplacian regularization can induce another type of sparsity (see the extra noise term $\xi_i$), which can not be ensured by the traditional Laplacian regularization. Moreover, the $p$-Laplacian regularization \cite{ZS05} can also be regarded as an ordinary $L_1$-generalization of the Laplacian regularization when $p=1$. According to \cite{CLK10}, by defining a matrix $C_p \in R^{\frac{k(k-1)}{2}\times k}$, the $p$-Laplacian regularization can be formulated as $||C_p \alpha_i||_1$, similar to our $L_1$-norm Laplacian regularization. Hence, we can similarly apply the $p$-Laplacian regularization with $p=1$ to constrained spare representation. However, such Laplacian regularization incurs large time cost due to the large matrix $C_p$ even for small neighborhood size (e.g. $k=90$).

Once we have constructed the $L_1$-graph $\mathcal{G}_{\mathcal{X}} = \{\mathcal{X}, W_{\mathcal{X}}\}$ over $\mathcal{X}$, we can similarly construct the $L_1$-graph $\mathcal{G}_{\mathcal{Y}}=\{\mathcal{Y}, W_{\mathcal{Y}}\}$ over $\mathcal{Y}$. Based on the two weight matrices, our inter-view constraint propagation can be performed with constrained graph construction (CGC) (as shown in Fig.~\ref{Fig.1})  using constrained sparse representation  (CSR) developed here.

\section{Application to Cross-View Retrieval}
\label{sect:cmr}

\begin{figure}[t]
\vspace{0.06in}
\begin{center}
\includegraphics[width=0.48\textwidth]{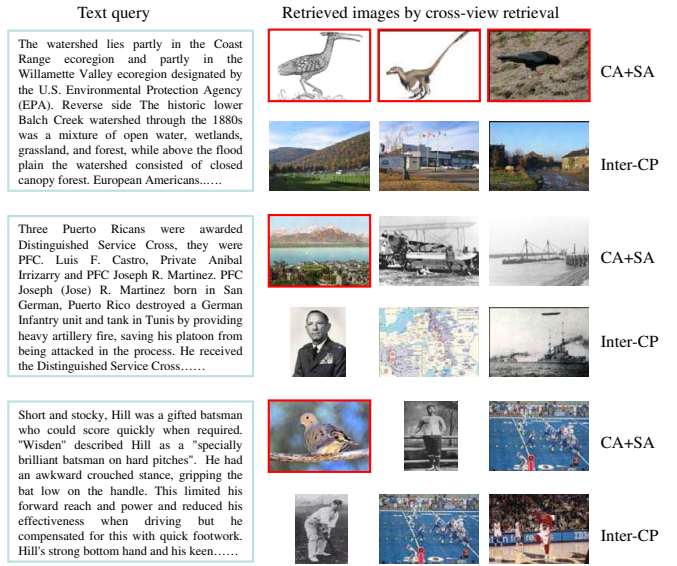}
\end{center}
\vspace{-0.10in} \caption{Cross-view retrieval examples on the
Wikipedia benchmark dataset \cite{RCC10}. Here, the incorrectly retrieved images are marked with red boxes. } \label{Fig.3} \vspace{-0.10in}
\end{figure}

When multiple views refer to text, image, audio and so on (see
Fig.~\ref{Fig.3}), the output of our inter-view constraint propagation actually can be viewed as the correlation between different media views. As we have mentioned, given the output $F^*=\{f^*_{ij}\}_{N \times M}$ of our inter-view
constraint propagation, $(x_i,y_j)$ denotes a must-link (or
cannot-link) constraint if $f^*_{ij}>0$ (or $<0$). Considering the
inherent meanings of must-link and cannot-link constraints, we can
state that: $x_i$ and $y_j$ are ``positively correlated" if
$f^*_{ij}>0$, while they are ``negatively correlated" if
$f^*_{ij}<0$. Hence, we can view $f^*_{ij}$ as the correlation
coefficient between $x_i$ and $y_j$. The distinct advantage of such
interpretation of $F^*$ as a correlation measure is that $F^*$ can
thus be used for ranking on $\mathcal{Y}$ given a query $x_i$ or
ranking on $\mathcal{X}$ given a query $y_j$. In fact, this is just
the goal of cross-view retrieval which has drawn much attention
recently \cite{RCC10}. That is, such task can be directly handled by
our inter-view constraint propagation.

In this paper, we focus on a special case of cross-view retrieval,
i.e. only text and image views are considered. In this case,
cross-view retrieval is somewhat similar to automatic image
annotation \cite{LW03,FML04,lu2011contextual,lu2011spatial} and image caption generation
\cite{FHS10,KPD11,OKB12}, since these three tasks all aim to learn
the relations between the text and image views. However, even if only
text and image views are considered, cross-view retrieval is still
quite different from automatic image annotation and image caption
generation. More concretely, automatic image annotation relies on
very limited types of textual representations and mainly associates
images only with textual keywords, while cross-view retrieval is
designed to deal with much more richly annotated data, motivated by
the ongoing explosion of Web-based content such as news
archives and Wikipedia pages. Similar to cross-view retrieval, image
caption generation can also deal with more richly annotated data
(i.e. captions) with respect to the textual keywords concerned in
automatic image annotation. However, this task tends to
model image captions as sentences by exploiting certain prior
knowledge (e.g. the $<$object, action, scene$>$ triplets used in
\cite{FHS10}), different from cross-view retrieval that focuses
on associating images with complete text articles using no prior
knowledge from the text view (any general textual representations are
applicable actually once their similarities are provided).

\begin{figure*}[t]
\vspace{0.06in} \centering
\begin{minipage}{0.23\textwidth}
\centering
\includegraphics[width=0.99\textwidth]{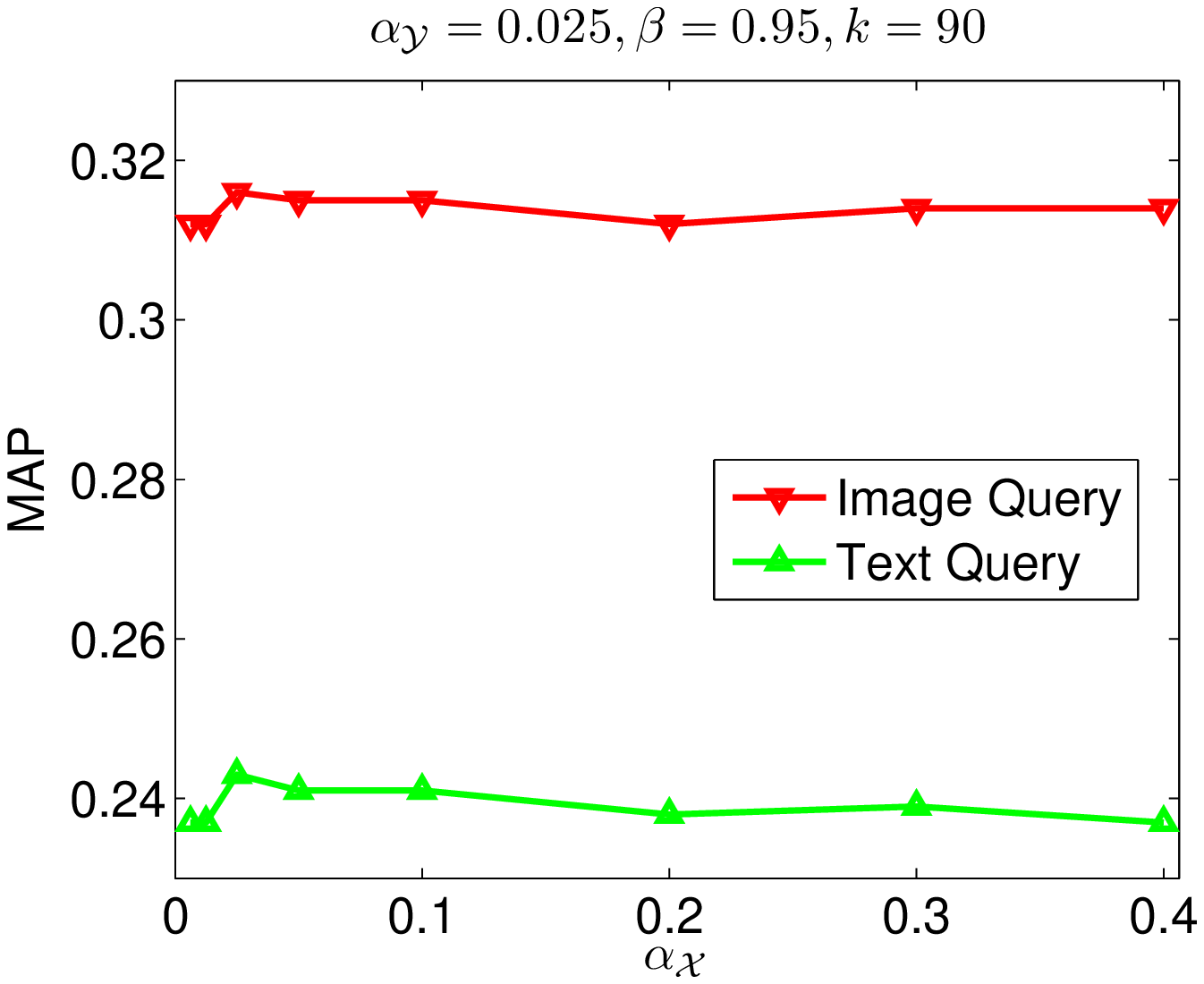}
\end{minipage}
\hspace{0.07in}
\begin{minipage}{0.23\textwidth}
\centering
\includegraphics[width=0.99\textwidth]{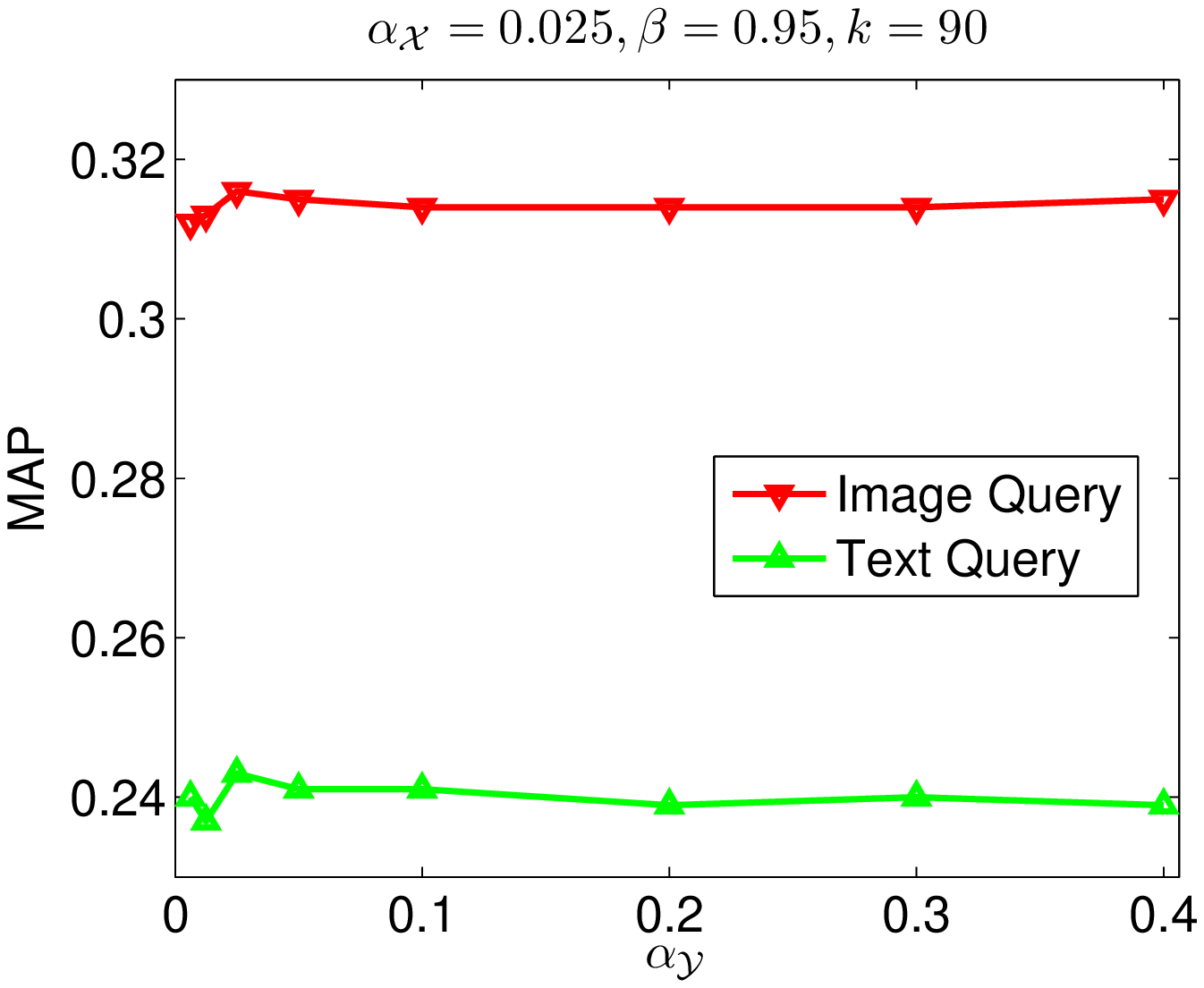}
\end{minipage}
\hspace{0.07in}
\begin{minipage}{0.23\textwidth}
\centering
\includegraphics[width=0.99\textwidth]{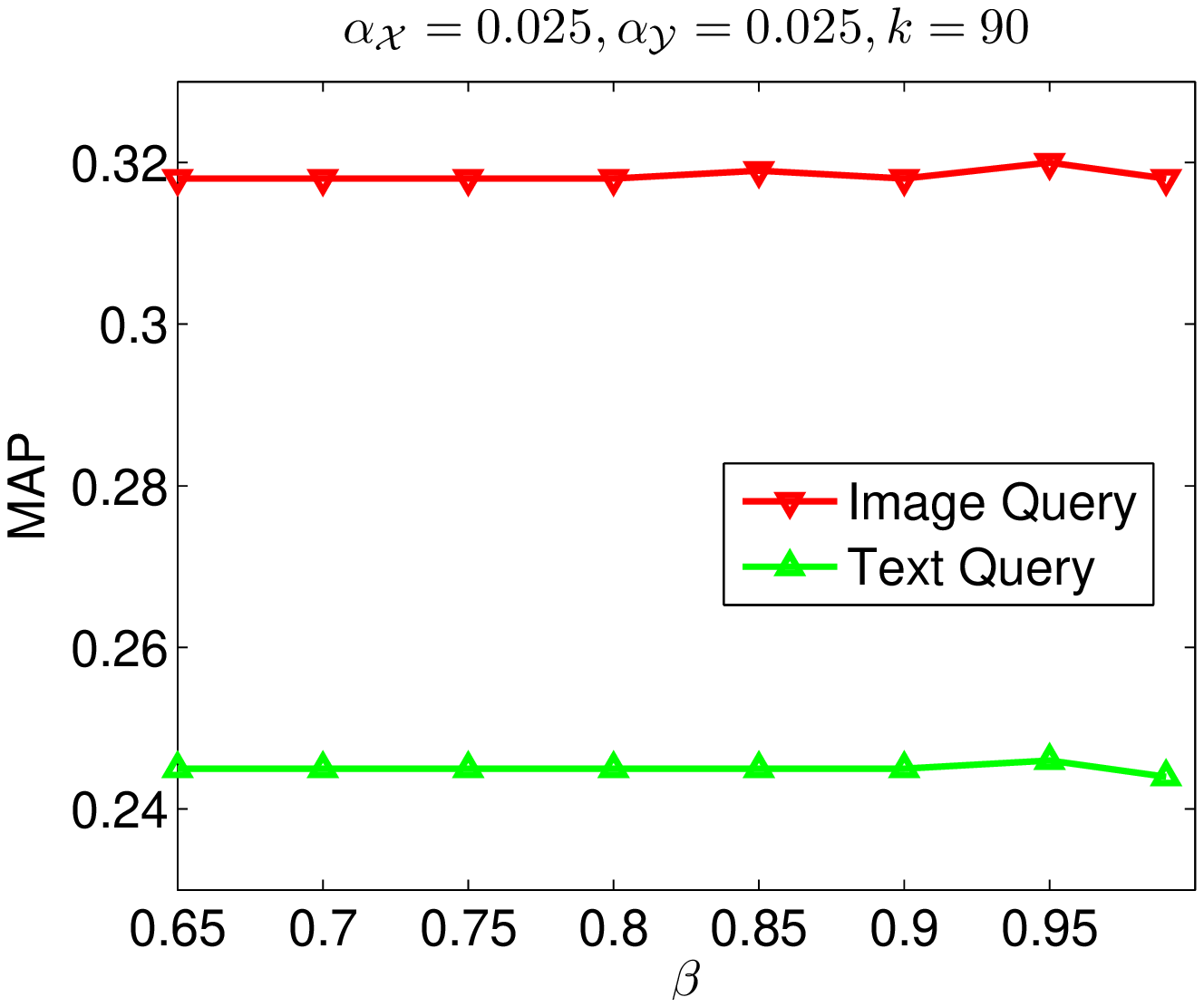}
\end{minipage}
\hspace{0.07in}
\begin{minipage}{0.23\textwidth}
\centering
\includegraphics[width=0.99\textwidth]{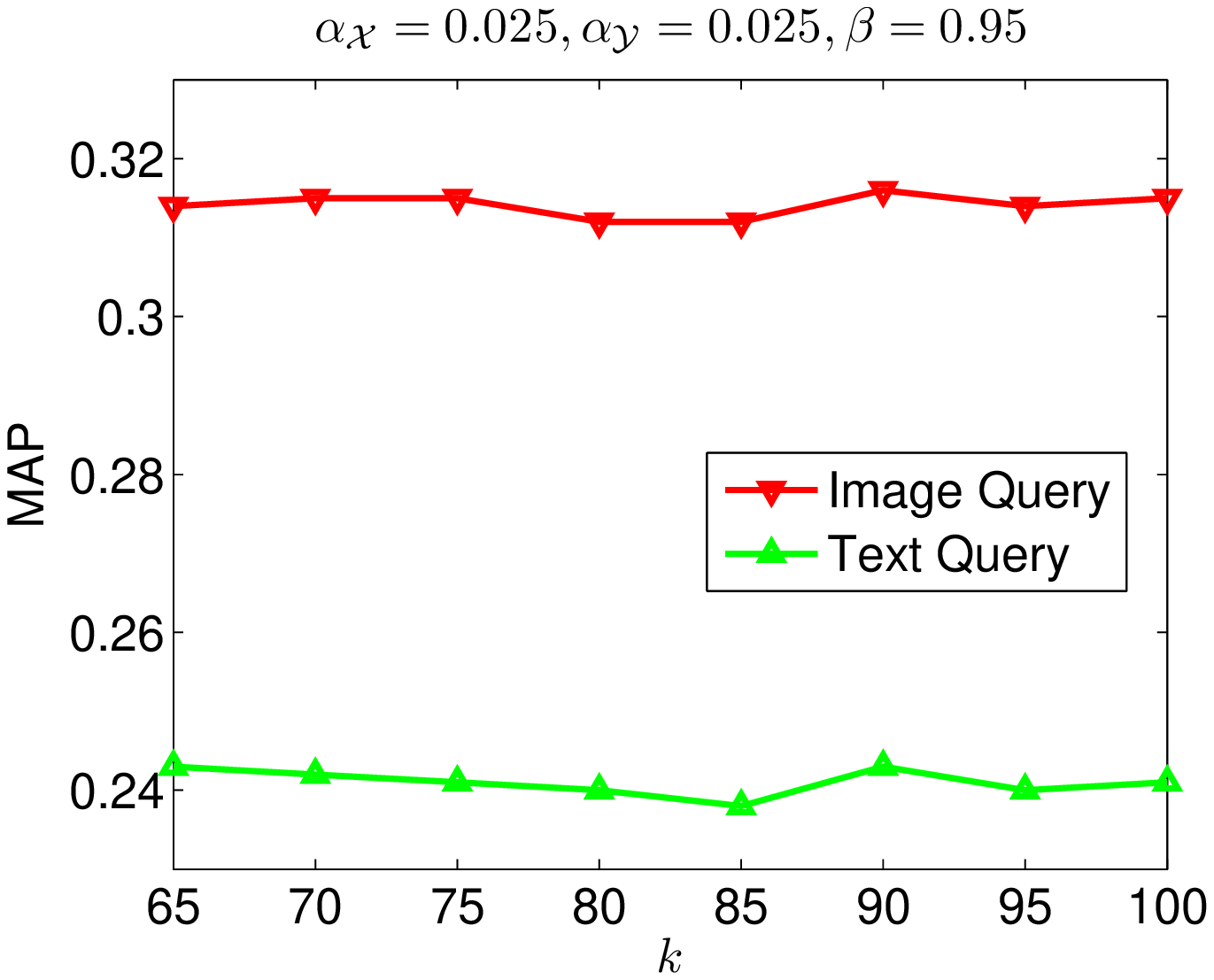}
\end{minipage}
\vspace{-0.02in} \caption{The cross-view retrieval results by cross-validation on the training set of the Wikipedia dataset for our Inter-CP algorithm (CSR is used here). } \label{Fig.4} \vspace{-0.15in}
\end{figure*}

In the context of cross-view retrieval, one notable recent work is
\cite{RCC10} which first learns the correlation between the text and image views with canonical correlation analysis (CCA) \cite{Hotelling36} and then achieves the abstraction by representing text and image at a more general semantic level. However, two separate steps, i.e. correlation analysis (CA) and semantic abstraction (SA), are involved in this modeling, and the use of semantic abstraction after CCA (i.e. CA+SA) seems rather ad hoc. Fortunately, this problem can be completely addressed by our inter-view constraint propagation (Inter-CP). The semantic information (e.g. class labels) associated with images and text can be used to define the initial must-link and cannot-link constraints based on the training dataset, while the correlation between text and image views can be explicitly learnt by the proposed algorithm in Section~\ref{sect:intercp}. That is, the correlation analysis and semantic abstraction has been successfully integrated in our inter-view constraint propagation framework. The effectiveness of such integration as compared to CA+SA \cite{RCC10} is preliminarily verified by several cross-view retrieval examples shown in Fig.~\ref{Fig.3}. Further verification will be provided in our later experiments. More notably, although only tested in cross-view retrieval, our inter-view constraint propagation can be readily extended to other multi-view tasks, since it has actually learnt the correlation between different views.

\section{Experimental Results}
\label{sect:exp}

In this section, our inter-view constraint propagation (Inter-CP) algorithm is evaluated in the challenging application of cross-view retrieval. We focus on comparing our Inter-CP algorithm with the state-of-the-art approach \cite{RCC10}, since they both consider not only correlation analysis (CA) but also semantic abstraction (SA) for text and image views. Moreover, we also make comparison with another two closely related approaches that integrate CA and SA for cross-view retrieval similar to \cite{RCC10} but perform correlation analysis by partial least squares (PLS) \cite{Wold85} and cross-modal factor analysis (CFA) \cite{LDL03} instead of CCA, respectively. In the following, these two CA+SA approaches are denoted as CA+SA (PLS) and CA+SA (CFA), while the state-of-the-art approach \cite{RCC10} is denoted as CA+SA (CCA). Finally, to show the effectiveness of constrained graph construction, we construct four types of graphs for our Inter-CP algorithm: $k$-NN graph ($k$-NN), $L_1$-graph using sparse representation (SR), $k$-NN graph using constrained weight adjustment (CWA), and $L_1$-graph using constrained sparse representation (CSR).

\subsection{Experimental Setup}

We select two different datasets for performance evaluation. The
first one is a Wikipedia benchmark dataset \cite{RCC10}, which
contains a total of 2,866 documents derived from Wikipedia's
``featured articles". Each document is actually a text-image pair,
annotated with a label from the vocabulary of 10 semantic classes.
This benchmark dataset \cite{RCC10} is split into a training set of
2,173 documents and a test set of 693 documents. Moreover, the second
dataset consists of totally 8,564 documents crawled from the photo
sharing website Flickr. The image and text views of each document
denote a photo and a set of tags provided by the users, respectively.
Although such text presentation does not take a free form as that for
the Wikipedia dataset, it is rather noisy since many of the tags may
be incorrectly annotated by the users. This Flickr dataset is
organized into 11 semantic classes. We split it into a training set
of 4,282 documents and a test set of the same size.

For the above two datasets, we take the same strategy as \cite{RCC10}
to generate both text and image representation. More concretely, in
the Wikipedia dataset, the text representation for each document is derived from a latent Dirichlet allocation model with 10 latent topics, while the image representation is based on a bag-of-words model with 128 visual words learnt from the extracted SIFT descriptors, just as \cite{RCC10}. Moreover, for the Flickr dataset, we generate the text and image representation similarly, and the main difference is that we select a relatively large visual vocabulary (of the size 2,000) for image representation and refine the noisy textual vocabulary to the size 1,000 by a preprocessing step for text representation.

In our experiments, the intra-view pairwise
constraints used for our CGC and inter-view pairwise
constraints used for our Inter-CP are initially derived from the class labels of the training documents of each dataset. The performance of our Inter-CP with CGC is evaluated on the test set. Here, two tasks of cross-view retrieval are considered: text retrieval using an image query, and image retrieval using a text query. In the following,
these two tasks are denoted as ``Image Query" and ``Text Query",
respectively. For each task, the retrieval results are measured with
mean average precision (MAP) which has been widely used in the image
retrieval literature \cite{GVS10}.

Let $\mathcal{X}$ denote the text representation and $\mathcal{Y}$
denote the image representation. For our Inter-CP
algorithm, we perform CGC over $\mathcal{X}$ and $\mathcal{Y}$ with the same $k$. The parameters of our Inter-CP algorithm with CGC can be selected by fivefold cross-validation on the training set. For example, according to Fig.~\ref{Fig.4}, we set the parameters of our Inter-CP (CSR is used for CGC) on the Wikipedia dataset as:
$\alpha_\mathcal{X}=0.025$, $\alpha_\mathcal{Y}=0.025$, $\beta=0.95$,
and $k=90$. It is noteworthy that our Inter-CP with CSR is not sensitive to these parameters. Moreover, the parameters of our Inter-CP with CWA can be similarly set to their respective optimal values. To
summarize, we have selected the best values for all the parameters of
our UCP algorithm with CGC by cross-validation on the training set. For fair comparison, we take the same parameter selection strategy for other closely related algorithms.

\begin{table}[t]
\vspace{0.06in} \caption{The cross-view retrieval results on the test set of the Wikipedia dataset measured by the MAP scores.} \label{Table.1}
\vspace{-0.15in}
\begin{center}
\tabcolsep0.18cm
\begin{tabular}{|c|ccc|}
\hline
Methods & Image Query  & Text Query & Average \\
\hline
CA+SA (PLS)      &  0.250   &  0.190  &  0.220 \\
CA+SA (CFA)      &  0.272   &  0.221  &  0.247 \\
CA+SA (CCA)      &  0.277   &  0.226  &  0.252 \\
Inter-CP+$k$-NN  &  0.329   &  0.256  &  0.293 \\
Inter-CP+SR      &  0.336   &  0.259  &  0.298 \\
Inter-CP+CWA     &  0.337   &  0.260  &  0.299 \\
Inter-CP+CSR     & \textbf{0.343} & \textbf{0.268} & \textbf{0.306} \\
\hline
\end{tabular}
\end{center}
\vspace{-0.15in}
\end{table}

\subsection{Retrieval Results}

The cross-view retrieval results on the two datasets are listed in
Tables~\ref{Table.1} and \ref{Table.2}, respectively. The immediate
observation is that we can achieve the best results when both
intra-view and inter-view pairwise constraints are exploited by
Inter-CP+CWA (or Inter-CP+CSR). This means that our Inter-CP with CGC can most effectively exploit the initial supervisory information provided for cross-view retrieval. As compared to the three CA+SA approaches by semantic abstraction after correlation analysis (via PLS, CFA, or CCA), our Inter-CP can seamlessly integrate these two separate steps and then lead to much better results. Moreover, the effectiveness of our CGC is verified by the comparison Inter-CP+CWA vs. Inter-CP+$k$-NN (or Inter-CP+CSR vs. Inter-CP+SR), especially on the Flickr dataset. As for our two CGC methods, CSR is shown to perform better than CWA, which is mainly due to the noise-robustness property of sparse representation.

It should be noted that our Inter-CP algorithm can be considered to provide an efficient solution, since it has a time cost proportional to the number of all possible pairwise constraints. This is also verified by our observations in the experiments. For example, the running time taken by CA+SA (CCA, CFA or PLS), Inter-CP+$k$-NN, and Inter-CP+CWA
on the Wikipedia dataset is 10, 24, and 55 seconds, respectively.
Here, we run all the algorithms (Matlab code) on a computer with 3GHz
CPU and 32GB RAM. Since our Inter-CP with CGC leads to
significantly better results, we prefer it to CA+SA in practice, regardless of its relatively larger time cost.

\begin{table}[t]
\vspace{0.06in} \caption{The cross-view retrieval results on the test set of the Flickr dataset measured by the MAP scores.} \label{Table.2}
\vspace{-0.15in}
\begin{center}
\tabcolsep0.18cm
\begin{tabular}{|c|ccc|}
\hline
Methods & Image Query  & Text Query & Average \\
\hline
CA+SA (PLS)      &  0.201  &  0.168  &  0.185 \\
CA+SA (CFA)      &  0.252  &  0.231  &  0.242 \\
CA+SA (CCA)      &  0.280  &  0.263  &  0.272 \\
Inter-CP+$k$-NN  &  0.495  &  0.483  &  0.489 \\
Inter-CP+SR      &  0.509  &  0.496  &  0.503 \\
Inter-CP+CWA     &  0.521  &  0.499  &  0.510 \\
Inter-CP+CSR     & \textbf{0.521} & \textbf{0.505} & \textbf{0.513}\\
\hline
\end{tabular}
\end{center}
\vspace{-0.15in}
\end{table}

\section{Conclusions}
\label{sect:con}

In this paper, we have investigated the challenging problem of pairwise constraint propagation on multi-view data. By decomposing the inter-view constraint propagation problem into a set of independent semi-supervised learning subproblems, we have uniformly formulated them as minimizing a regularized energy functional. More importantly, these semi-supervised learning subproblems can be solved efficiently using label propagation with $k$-NN graph. We then develop two constrained graph construction methods for our inter-view constraint propagation, and the obtained two graphs can be considered as the variants of $k$-NN graph. The experimental results in cross-view retrieval have shown the promising performance of our inter-view constraint propagation with constrained graph construction. For future work, our method will be extended to other multi-view tasks.

\section*{Acknowledgements}

This work was supported by National Natural Science Foundation of
China under Grants 61202231 and 61222307, National Key Basic Research Program (973 Program) of China under Grant 2014CB340403, Beijing Natural Science Foundation of China under Grant 4132037, the Fundamental Research Funds for the Central Universities and the Research Funds of Renmin University of China under Grant 14XNLF04, and a grant from Microsoft Research Asia.

\IEEEtriggeratref{34}
%\bibliographystyle{IEEEtran}
%\bibliography{mpcp}


\end{document}